\documentclass[preprint,12pt]{elsarticle}
\usepackage[T1]{fontenc}
\usepackage{amssymb}
\usepackage{amsmath}
\usepackage{lineno}
\usepackage{adjustbox}
\usepackage{booktabs}
\usepackage[hyphens]{url}
\usepackage[unicode,hidelinks]{hyperref} 
\usepackage{comment}
\usepackage{multirow}
\usepackage{siunitx}
\journal{Expert Systems with Applications}

\begin{document}

\begin{frontmatter}


\title{RatioWaveNet: A Learnable RDWT Front-End for Robust and Interpretable EEG Motor-Imagery Classification}

 \author[unict,unipa]{Marco Siino\corref{cor1}}
 \ead{marco.siino@unict.it, marco.siino@unipa.it}
 \cortext[cor1]{Corresponding author.}
 \author[unipa]{Giuseppe Bonomo}
 \author[unipa]{Rosario Sorbello}
 \author[unipa]{Ilenia Tinnirello}
 \affiliation[unict]{organization={Department of Electrical, Electronics and Informatics Engineering, University of Catania},
             city={Catania},
             postcode={95131},
             country={Italy}}
 \affiliation[unipa]{organization={Department of Engineering, University of Palermo},
             city={Palermo},
             postcode={90128},
             country={Italy}}

\begin{abstract}
Brain–computer interfaces (BCIs) based on motor imagery (MI) translate covert movement intentions into actionable commands, yet reliable decoding from non-invasive EEG remains challenging due to nonstationarity, low SNR, and subject variability. We present \textit{RatioWaveNet}, which augments a strong temporal CNN–Transformer backbone (i.e., \textit{ TCFormer}) with a trainable, Rationally-Dilated Wavelet Transform (RDWT) front end. The RDWT performs an undecimated, multi-resolution subband decomposition that preserves temporal length and shift-invariance, enhancing sensorimotor rhythms while mitigating jitter and mild artifacts; subbands are fused via lightweight grouped 1-D convolutions and passed to a multi-kernel CNN for local temporal–spatial feature extraction, a grouped-query attention encoder for long-range context, and a compact TCN head for causal temporal integration. 

Our goal is to test whether this principled wavelet front end improves robustness precisely where BCIs typically fail—on the hardest subjects—and whether such gains persist on average across seeds under both intra- and inter-subject protocols. On BCI-IV-2a and BCI-IV-2b, across five seeds, RatioWaveNet improves \emph{worst-subject} accuracy over the Transformer backbone by \textbf{+0.17} / \textbf{+0.42} percentage points (Sub-Dependent / LOSO) on 2a and by \textbf{+1.07} / \textbf{+2.54} percentage points on 2b, with consistent average-case gains and modest computational overhead. These results indicate that a simple, trainable wavelet front end is an effective plug-in to strengthen Transformer-based BCIs, improving worst-case reliability without sacrificing efficiency. 


\end{abstract}


\end{frontmatter}


\section{Introduction}
\label{sec:Introduction}

Brain–computer interfaces (BCIs) establish a direct communication pathway between neural activity and external devices by decoding brain signals into actionable commands, with the potential to transform both clinical rehabilitation and human-computer interactions and thereby improve quality of life \cite{altaheri2023_review,altaheri2025_foundation}. Among brain-sensing modalities, electroencephalography (EEG) is particularly well suited to real-world deployment because it is non-invasive, inexpensive, portable, and offers millisecond-level temporal resolution; these properties have underpinned a broad spectrum of EEG-based applications spanning cognitive skill assessment \cite{lin2024_expertise}, driver vigilance estimation \cite{tang2024_hmstene}, emotion recognition \cite{gao2025_multidomain}, and human–robot interaction \cite{alquraishi2018_exoskeletons}. In healthcare, EEG is a key enabler of smart health ecosystems \cite{clark2024_smarthealth}, supporting tasks such as automated sleep staging \cite{masad2024_sleep}, seizure detection and monitoring \cite{alhussein2018_seizure}, and neurorehabilitation after stroke \cite{lopezlarraz2018_bmi_rehab}. Within this context, Motor Imagery (MI) - the covert rehearsal of movement without execution - has become one of the most widely adopted BCI paradigms \cite{altaheri2023_review}. MI-EEG systems have been leveraged in medical scenarios including post-stroke motor rehabilitation \cite{lopezlarraz2018_bmi_rehab,cantillo2018_orthosis}, control of prosthetic and exoskeletal devices \cite{alquraishi2018_exoskeletons} brain-controlled wheelchair navigation \cite{fernandez2016_wheelchairs,li2013_wheelchair}, and thought-to-text communication \cite{zhang2018_percom_braintyping}. Beyond clinical use, MI-EEG has enabled non-medical applications such as drone piloting \cite{lafleur2013_quadcopter}, brain-actuated vehicle control \cite{yu2016_braincar}, intent recognition for smart-home environments \cite{zhang2017_smartliving}, and interactive virtual/augmented experiences \cite{djamal2017_game,ang2012_fbcsp}. Notwithstanding these advances, translating MI-based BCIs to unconstrained settings remains difficult due to the low signal-to-noise ratio and pronounced non-stationarity of EEG, together with substantial inter- and intra-subject variability; collectively, these factors hinder the extraction of robust, discriminative neural patterns and limit generalization performance across users and sessions\cite{altaheri2023_review}.

Decoding MI from EEG is intrinsically difficult: the signals exhibit a low signal-to-noise ratio, marked non-stationarity, and substantial variability both across and within subjects, all of which obscure the task-related neural patterns. Robust, discriminative representations are therefore hard to extract, and models must generalise across sessions and users despite drift, artifacts, and heterogeneous spectral content. These challenges motivate the development of approaches that are not only accurate but also resilient and broadly generalisable.

A wide spectrum of machine learning (ML) and deep learning (DL) methods has been explored to meet these requirements. Among traditional ML pipelines based on hand-crafted features, filter-bank spatial filtering - most notably the Filter-Bank Common Spatial Pattern (FBCSP)-has shown consistently competitive performance by decomposing the EEG into multiple frequency sub-bands and deriving discriminative spatial–spectral projections \cite{ang2012_fbcsp}. In contrast, modern DL models learn features directly from minimally processed EEG, reducing the reliance on manual engineering and enabling the extraction of richer hierarchical representations that better capture the interplay of spatial, spectral, and temporal structure. DL has rapidly advanced MI-EEG classification, spawning a diverse family of neural architectures that learn representations directly from the signal. Prominent lines of work include Convolutional Neural Networks (CNNs) \cite{altuwaijri2022_multibranch_se,amin2021_i2mtc_att_inception,amin2022_tii_inception_lstm,lawhern2018_eegnet,li2020_tnsre_fusion}, autoencoders \cite{hassanpour2019_expertsyst}, recurrent models such as RNN/LSTM/GRU \cite{kumar2021_opticalplus,luo2018_bmc_rnn}, temporal convolutional networks (TCNs) \cite{ingolfsson2020_eegtcnet,musallam2021_tcn_fusion,altaheri2023_physattn_tcn,altaheri2023_dynamic_conv_attn}, deep belief networks \cite{xu2020_sensors_multiview}, and, more recently, Transformer-based approaches \cite{altaheri2023_physattn_tcn,altaheri2023_dynamic_conv_attn,zhao2025_mscformer,zhao2024_ctnet,song2023_eegconformer}. Among these, CNNs have emerged as the default backbone for MI-EEG because they can distill hierarchical spatiotemporal structure from minimally processed inputs \cite{altaheri2023_review}. Both compact designs \cite{lawhern2018_eegnet} and deeper variants \cite{li2019_ieeeaccess_imaging} have been explored, alongside numerous derivatives—inception-style \cite{amin2021_i2mtc_att_inception,amin2022_tii_inception_lstm}, residual and 3D extensions \cite{liu2021_brainsci_densely3d}, multi-scale \cite{li2020_tnsre_fusion}, multi-branch \cite{altuwaijri2022_multibranch_se,liu2021_brainsci_densely3d}, and attention-augmented CNNs \cite{altuwaijri2022_multibranch_se,amin2021_i2mtc_att_inception,amin2022_tii_inception_lstm,li2020_tnsre_fusion,altaheri2023_dynamic_conv_attn,zhang2020_IEEEJBHI_graphatt,wimpff2024_jne_channelattn}. Canonical baselines include DeepConvNet and ShallowConvNet \cite{schirrmeister2017_hbm}, while EEGNet \cite{lawhern2018_eegnet} popularised depthwise–separable convolutions to factor temporal and spatial filtering efficiently; subsequent systems such as BaseNet \cite{wimpff2024_jne_channelattn} combined EEGNet/ShallowConvNet priors with channel-attention modules. Despite their competitiveness, purely convolutional pipelines are bounded by finite receptive fields, which can underspecify long-range temporal dependencies in MI-EEG.

To capture sequential structure more explicitly, temporal modelling has leveraged both recurrence and dilated convolutions. Early work coupled hand-engineered filter-bank features with LSTM/GRU classifiers \cite{kumar2021_opticalplus,luo2018_bmc_rnn}. TCNs \cite{bai2018_empirical} provide a compelling alternative: stacks of causal, dilated convolutions expand the effective context while retaining stable optimisation and modest parameter counts, properties that have driven their adoption in MI-EEG \cite{ingolfsson2020_eegtcnet,musallam2021_tcn_fusion,altaheri2023_physattn_tcn,altaheri2023_dynamic_conv_attn}. Representative examples include EEG-TCNet, which integrates a TCN head with EEGNet features \cite{ingolfsson2020_eegtcnet}, multi-level fusion schemes that strengthen cross-scale aggregation \cite{musallam2021_tcn_fusion}, and attention-guided CNN–TCN hybrids achieving state-of-the-art results \cite{altaheri2023_physattn_tcn,altaheri2023_dynamic_conv_attn}. Complementary to these trends, architectures such as TCANet jointly exploit multi-scale CNN blocks, temporal convolutions, and self-attention to couple local and global cues within a lightweight design \cite{zhao2025_tcanet}. In parallel, attention mechanisms have been infused across model families to prioritise informative channels, time segments, and topologies, improving noise suppression and class separability in MI-EEG \cite{altuwaijri2022_multibranch_se,amin2021_i2mtc_att_inception,amin2022_tii_inception_lstm,li2020_tnsre_fusion,altaheri2023_physattn_tcn,zhang2020_IEEEJBHI_graphatt,wimpff2024_jne_channelattn}. Examples range from graph-aware models with LSTM and self-attention \cite{zhang2020_IEEEJBHI_graphatt}, to inception–LSTM hybrids with attention \cite{amin2021_i2mtc_att_inception,amin2022_tii_inception_lstm}, to multi-branch CNNs enhanced by squeeze-and-excitation blocks \cite{altuwaijri2022_multibranch_se}, and comparative frameworks that systematically assess channel-attention strategies \cite{wimpff2024_jne_channelattn}.

Building on this momentum, Transformer-based models bring global self-attention and efficient parallelism, traits that are attractive for long-sequence EEG analysis. Nonetheless, when applied directly to raw or lightly preprocessed EEG, Transformers can overfit small datasets and may lack the locality bias needed to capture short-range sensorimotor structure. Consequently, many recent systems deploy hybrid CNN→Transformer pipelines in which convolutions provide local feature extractors and the Transformer layers encode global context, followed by a classifier \cite{altaheri2023_physattn_tcn,altaheri2023_dynamic_conv_attn,zhao2025_mscformer,zhao2024_ctnet,song2023_eegconformer,xie2022_tnsre_transformer,sun2022_jbhi_ieeg_transformer,ding2024_jbhi_eegdeformer}. Notable instantiations include EEG Conformer \cite{song2023_eegconformer}, CTNet \cite{zhao2024_ctnet}, and MSCFormer \cite{zhao2025_mscformer}. Beyond this sequential hand-off, specialised hybrids tailor the Transformer block or the tokenisation scheme to EEG. For instance, ATCNet introduces sliding-window (patch) embeddings and replaces the standard MLP head with a temporal convolutional module, thereby reinforcing temporal dynamics within the Transformer stage and improving downstream MI decoding \cite{altaheri2023_physattn_tcn}.

We introduce \textbf{RatioWaveNet}, an extension of a TCFormer-style backbone that explicitly bridges global attention and causal temporal decoding. Upstream of the MK–CNN → Transformer → TCN backbone, RatioWaveNet applies an undecimated RDWT to raw EEG, producing scale-separated subbands that remain sample-aligned. The processing stream remains compact and purposeful: a multi-kernel CNN (MK--CNN) first extracts short-range temporal--spectral primitives using parallel 1-D filters targeted to distinct EEG bands; a Transformer with grouped-query attention (GQA) and rotary positional embeddings (RoPE) then assembles these primitives into a time-indexed latent sequence that captures long-range dependencies at favorable compute/memory cost. This placement is deliberate: purely attention-driven features can diffuse band-limited motor-imagery cues across channels and time, whereas undecimated, rational dilations re-express them as temporally aligned subbands that are near shift-invariant and more finely tuned than integer-rate designs around the $\mu/\beta$ rhythms and their harmonics. Subbands are fused by lightweight grouped $1{\times}k$ convolutions (depthwise--separable by default) with an optional squeeze-and-excite gate that reweights scales prior to causal decoding; the grouped TCN then performs dilated causal integration with receptive fields chosen to match the effective horizons implied by the rational dilations, enabling coherent aggregation across scales while preserving strict causality for online settings. Because the RDWT is implemented with 1-D convolutions and no decimation, its cost scales linearly with sequence length and number of scales. For online use, causal padding can be employed. GQA curbs the quadratic overhead of self-attention and the MK--CNN maintains parameter efficiency through parallel but shallow kernels. Filters in the RDWT stage can be instantiated from analytic wavelet prototypes to retain interpretability, or made lightly trainable under norm constraints to adapt passbands to subject- or dataset-specific spectra; in both cases, the grouped fusion preserves scale identity up to the TCN, acting as a structure-preserving adapter between global attention and causal decoding. In essence, RatioWaveNet couples rich local feature extraction, efficient global context modeling, explicit multi-scale regularization, and long-range causal integration in a single end-to-end pipeline aligned with the band-limited, mildly non-stationary nature of MI-EEG.

\begin{figure}[h]
    \centering    \includegraphics[width=1.0\linewidth]{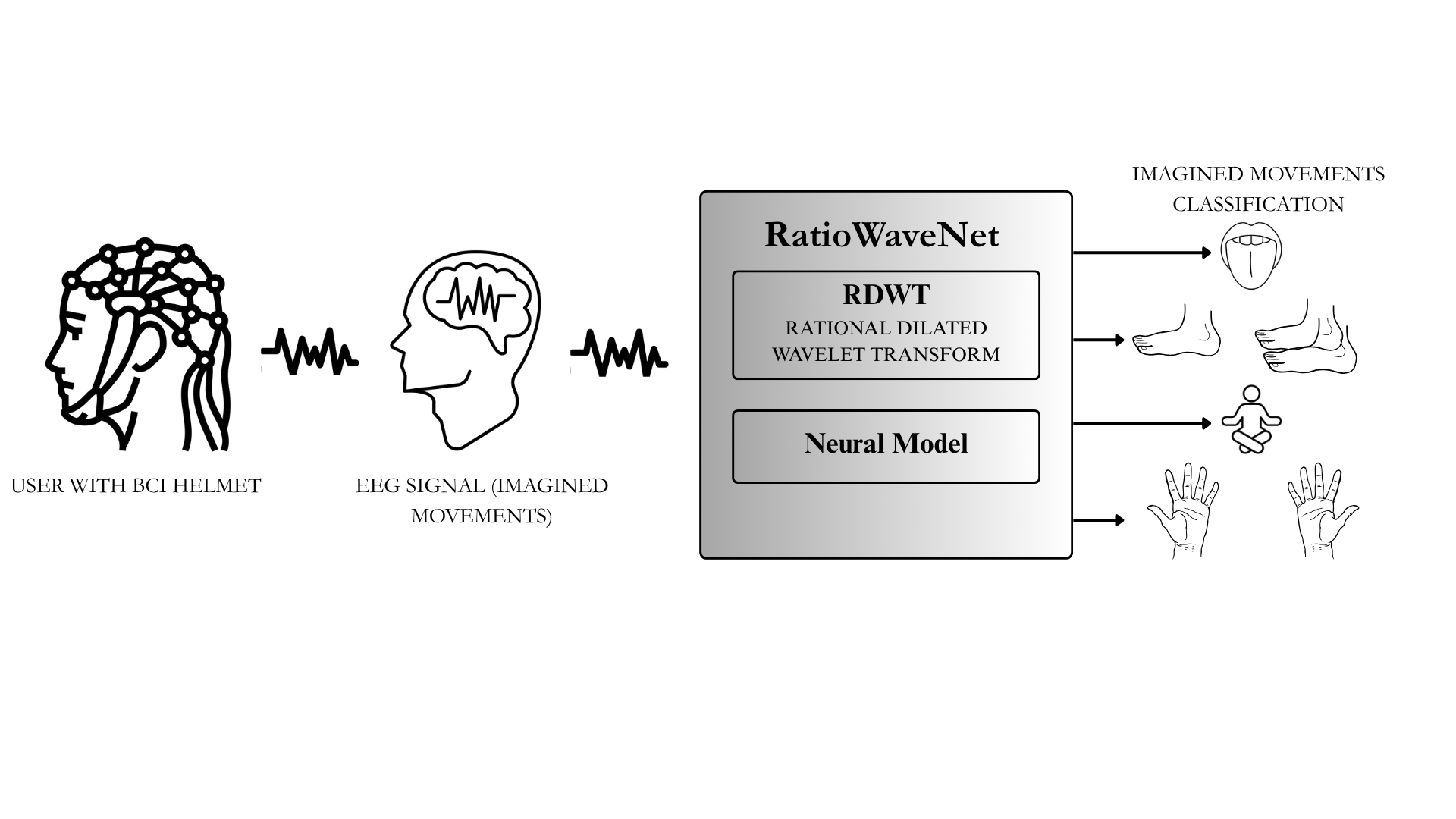}
    \caption{Overview of the proposed framework: EEG signals are processed via RatioWaveNet (RDWT + Neural Model) for imagined movements classification.}
    \label{fig:eeg_schema}
\end{figure}

In summary, the main contributions of this work are:

\begin{itemize}
    \item We introduce a \textbf{trainable, shift-invariant RDWT front end} for MI-EEG: rational scale parameters learned via a tempered logistic map; lightly trainable db4 prototypes resampled at the learned scales.
    \item We devise a \textbf{parallel RDWT ensemble} and \textbf{learned softmax fusion}, complemented by branch-dropout and small scale-logit jitter.
    \item We employ an \textbf{efficient attention block with relative/rotary positional phase} to capture long-range temporal dependencies at reduced memory/compute, and a \textbf{lightweight TCN head} with dilated causal convolutions and grouped residuals to consolidate mid-/long-range dynamics.
    \item On robustness-oriented evaluation, \textbf{RatioWaveNet} maintains or modestly improves worst-case performance over a strong transformer baseline with limited overhead.
\end{itemize}

This work delivers a principled, modular pipeline for EEG decoding in which adaptive wavelet-domain preprocessing and compact CNN–attention–TCN blocks jointly improve \emph{worst-case} subject robustness while preserving computational efficiency.

\section{Related Work}

\subsection{Time-Frequency Analysis in EEG}

Traditional signal processing techniques have long served as the foundation of EEG-based BCIs, offering a means to extract frequency-domain features from inherently non-stationary neural signals. The STFT is widely used to obtain time-localized spectral information via a sliding-window Fourier transform. However, its reliance on a fixed window size inherently limits the trade-off between time and frequency resolution \cite{vidyasagar2024signal}, making it less suited for tracking rapid transients in EEG.

Wavelet-based approaches, particularly the DWT, have demonstrated improved performance for EEG decoding by enabling multi-resolution analysis of signal components. DWT is particularly effective in isolating oscillatory activity across characteristic EEG frequency bands \cite{zhou2020wavelet}, and has been used both as a standalone preprocessing step and in conjunction with machine learning classifiers.

Recent advances have combined DWT with deep learning architectures, such as Convolutional Neural Networks (CNNs) and Recurrent Neural Networks (RNNs), where wavelet-decomposed EEG signals serve as inputs to enhance discriminative feature extraction \cite{bashivan2015learning}. Despite these improvements, both STFT and conventional DWT employ fixed or integer-valued dilation/scaling factors, which can fail to capture the variable temporal dynamics and scale-dependent structures present in EEG recordings.

To address this limitation, the RDWT was introduced as a more flexible alternative \cite{bayram2009rdwt}. RDWT uses non-integer dilation factors, such as $3/2$ or $5/3$, to adapt the frequency resolution to the underlying signal structure more effectively. This provides a better time-frequency localization for non-stationary signals, making it particularly suitable for EEG preprocessing in BCI applications.

\subsection{Deep Learning Architectures for EEG Decoding}

The integration of deep learning models has substantially advanced EEG-based classification tasks, particularly in motor imagery and event-related potential decoding. Among early efforts, EEGNet \cite{lawhern2018_eegnet} introduced a lightweight CNN architecture leveraging depthwise and separable convolutions to efficiently capture temporal and spatial features, setting a strong baseline across multiple BCI benchmarks.

ShallowConvNet \cite{schirrmeister2017_hbm} further simplified the architecture by employing shallow temporal and spatial convolutions followed by mean pooling, balancing performance and interpretability. More recent works, such as MBEEG\_SENet \cite{altuwaijri2022_multibranch_se}, introduced multi-branch pipelines combined with Squeeze-and-Excitation (SE) modules, enhancing the extraction of multiscale features by dynamically reweighting channel-wise activations.

TCNs have also shown strong promise in capturing long-range temporal dependencies within EEG sequences. For example, EEG-TCNet \cite{ingolfsson2020_eegtcnet} combined the EEGNet backbone with dilated TCN layers to model extended temporal contexts while preserving compactness. Similarly, ATCNet \cite{altaheri2023_physattn_tcn} integrated MHA with a TCN backbone, offering robust modeling of both short- and long-term patterns in EEG signals.

Despite these advances, the majority of deep learning models operate directly on raw or uniformly preprocessed EEG inputs, without incorporating adaptive signal transformation stages. This limits their ability to generalize across subjects and sessions with different signal artifacts or temporal variations.

RatioWaveNet couples a rational dilated wavelet transform (RDWT) front-end with the TCFormer backbone to jointly exploit multi-resolution spectral cues and global temporal dependencies in EEG. The RDWT module decomposes the raw signal into shift-invariant sub-bands and performs light denoising, enhancing band-specific transients while preserving temporal locality. These sub-band features are fused and passed to TCFormer, whose multi-kernel CNN (MK-CNN) expands representational capacity via parallel temporal kernels aligned to distinct EEG frequency ranges; a subsequent dimensionality-reduction step and group-wise attention emphasize the most informative scales. A grouped-query Transformer with rotary positional embeddings (RoPE) then models long-range temporal structure at reduced memory and compute cost, and a TCN head performs the final temporal decoding. This RDWT-TCFormer integration yields representations that are simultaneously noise-robust and temporally discriminative, enabling efficient and accurate MI-EEG classification suitable for practical BCI use.


\section{Proposed Method}

We introduce \textit{RatioWaveNet}, an RDWT--TCFormer model for MI-EEG classification. RDWT supplies shift-invariant, denoised multi-resolution features, while \textit{TCFormer}\cite{Altaheri2025}-combining MK-CNN, group-wise attention, a grouped-query Transformer with RoPE, and a TCN head-learns band-specific patterns and long-range temporal structure efficiently. This design addresses nonstationarity, low SNR, and inter-session variability within a unified, computationally efficient pipeline.

\subsection{Wavelet-Domain Preprocessing}
Let $x_{e,c}(t)$ denote the EEG for event $e$, channel $c$, and time $t\!\in\!\{1,\dots,T\}$. We employ a trainable rational dilated wavelet transform (RDWT) as front end that preserves temporal length (no decimation) and thus shift-invariance, a desirable property under nonstationarity and low SNR. In our implementation the number of analysis levels is \emph{fixed to four} ($L{=}4$). Each level is governed by a learnable rational scale $s_l\!\in\![1,S_{\max}]$ ($S_{\max}{=}4$) obtained via a tempered logistic map
\begin{equation}
s_l \;=\; 1 + (S_{\max}-1)\,\sigma\!\left(\frac{z_l}{\kappa}\right),
\label{eq:scale-param}
\end{equation}
with free logits $z_l$, temperature $\kappa\!>\!0$, and initialization anchored to rational dilations $1.5,\,5/3,\,7/4,\,9/5$. Starting from Daubechies-4 prototypes $h_{\mathrm{low}},h_{\mathrm{high}}\!\in\!\mathbb{R}^{K_0}$ (trainable with small deviations), level-specific analysis filters are built by linear resampling to length $K$ and $\ell_1$ normalization,
\begin{equation}
\tilde h^{(l)}_{\mathrm{low}}=\mathcal{N}_1\{\mathcal{R}(h_{\mathrm{low}};s_l,K)\},\qquad
\tilde h^{(l)}_{\mathrm{high}}=\mathcal{N}_1\{\mathcal{R}(h_{\mathrm{high}};s_l,K)\},
\label{eq:resample}
\end{equation}
where $\mathcal{R}$ is a 1-D stretch/compress interpolation and $\mathcal{N}_1\{v\}=v/(\|v\|_1+\varepsilon)$. With “same” padding and depthwise application (groups$=C$), the per-level analysis reads
\begin{equation}
a^{(0)}_{e,c}(t)=x_{e,c}(t),\quad
a^{(l)}_{e,c}(t)=(a^{(l-1)}_{e,c}\!*\tilde h^{(l)}_{\mathrm{low}})(t),\quad
d^{(l)}_{e,c}(t)=(a^{(l-1)}_{e,c}\!*\tilde h^{(l)}_{\mathrm{high}})(t).
\label{eq:analysis}
\end{equation}
Detail coefficients are denoised through a \emph{trainable} soft-threshold per level,
\begin{equation}
\tilde d^{(l)}_{e,c}(t)=\operatorname{sign}\!\big(d^{(l)}_{e,c}(t)\big)\max\!\big(|d^{(l)}_{e,c}(t)|-\tau_l,0\big),\qquad
\tau_l=\operatorname{softplus}(\theta_l),
\label{eq:soft-thresh}
\end{equation}
and scaled by learned gains $\alpha_l$. During training we apply \emph{level-dropout} with probability $p_{\mathrm{lev}}$ (per-sample or per-batch) that stochastically disables entire bands, improving robustness. The synthesis is shift-invariant and preserves the $(B,C,T)$ shape,
\begin{equation}
y_{e,c}(t)=a^{(L)}_{e,c}(t)+\sum_{l=1}^{L} m^{(l)}\,\alpha_l\,\tilde d^{(l)}_{e,c}(t),
\label{eq:recon}
\end{equation}
where $m^{(l)}\!\in\!\{0,1\}$ is active only at training time. To stabilize scale learning and promote diversity across the four levels, we include a lightweight regularizer
\begin{equation}
\mathcal{L}_{\mathrm{reg}}=
\lambda_{\mathrm{bar}}\!\sum_{l=1}^{L}\!\big(e^{-k(s_l-1)}+e^{-k(S_{\max}-s_l)}\big)
+\lambda_{z}\|z-z_\mu\|_2^2
+\lambda_{\mathrm{spr}}\!\!\sum_{l\neq j}\!e^{-\gamma|s_l-s_j|},
\label{eq:reg}
\end{equation}
where the first term discourages boundary sticking, the second anchors $z$ to its rational-seed initialization $z_\mu$, and the third (\emph{spread loss}) separates scales to reduce redundancy. Finally, we support an \emph{optional hybrid fusion} that mixes the raw stream and the RDWT output through learned convex weights (or light channel-wise attention) while preserving tensor shape for downstream blocks:
\begin{equation}
[\beta_{\mathrm{raw}},\beta_{\mathrm{rdwt}}]=\operatorname{softmax}(\eta),\qquad
y^{\mathrm{hyb}}_{e,c}(t)=\beta_{\mathrm{raw}}\,x_{e,c}(t)+\beta_{\mathrm{rdwt}}\,y_{e,c}(t).
\end{equation}
Overall, the four-level trainable RDWT yields a denoised, shift-invariant, multi-resolution representation tailored to the data, providing informative inputs and improved robustness to nonstationarity and inter-trial variability.

\subsection{Convolutional Feature Extraction}

The convolutional block in RatioWaveNet is designed to extract both spatial and temporal features from the reconstructed multichannel EEG signal. The structure comprises three convolutional layers, including depthwise and temporal convolutions. The first layer performs temporal convolution with $F_1$ filters of size $(K_T, 1)$ to extract frequency-specific patterns. Subsequently, a depthwise convolution with filter size $(1, C)$ and depth multiplier $D$ captures spatial interactions across EEG channels. A second temporal convolution layer with $F_2 = F_1 \times D$ filters fuses the spatiotemporal features. Each convolutional operation is followed by batch normalization and ELU activation. Dropout and average pooling (with pool size $(8,1)$) are applied to improve generalization and reduce computational load.

\subsection{Multi-Head Attention-Based Encoding}

To enhance the model's ability to capture temporal dependencies, the intermediate features are segmented into overlapping windows and processed independently using a shared MHA mechanism. This allows the model to dynamically weight important time steps within each segment. Given query ($Q$), key ($K$), and value ($V$) matrices \cite{vaswani2017attention} obtained through learned linear projections, the MHA operation is defined as:

\begin{equation}
\text{Attention}(Q, K, V) = \text{softmax}\left(\frac{QK^\top}{\sqrt{d_k}}\right)V,
\label{eq:mha}
\end{equation}

\noindent where $d_k$ is the dimensionality of the key vectors. This module enables the model to jointly attend to multiple representation subspaces, improving the interpretability and discriminative power of the encoded EEG features.

\paragraph{Grouped-Query Attention (GQA) and RoPE.}
We adopt grouped-query attention to reduce memory by sharing key–value projections
across groups of query heads. Let $h_q$ be the number of query heads and $h_{kv}<h_q$
the number of shared key–value heads; queries are split into $G=h_q/h_{kv}$ groups,
each attending to the same $K,V$. We further employ rotary positional embeddings (RoPE),
injecting relative phase into $(Q,K)$ prior to attention, which stabilizes long-range
temporal modeling without adding parameters.

\subsection{Temporal Modeling - TCN}

The attention-enhanced embeddings are fed into a TCN block, which employs dilated causal convolutions with residual connections to model long-range dependencies. This design facilitates an exponentially growing receptive field while maintaining computational efficiency. Each TCN layer consists of Conv-BN-Activation submodules, where increasing dilation factors capture dependencies across different temporal scales. The TCN structure preserves causality and sequence order, making it well-suited for EEG decoding tasks. The final TCN output is passed through a dense layer and softmax activation to yield class probabilities over motor imagery tasks.

\subsection{Summary}

RatioWaveNet offers a modular and interpretable architecture that fuses rational wavelet-domain features, attention-based encoding, and deep temporal modeling. Its design is particularly suited for nonstationary, multiscale EEG data and achieves robust performance across subjects and sessions in motor imagery classification tasks.

\section{Experimental Setup}

The RatioWaveNet model was implemented and evaluated using the PyTorch with CUDA on a single GPU. All experiments were conducted on a workstation equipped with an Intel\textregistered\ Xeon W7-3455 processor, an NVIDIA RTX 6000 Ada Generation GPU featuring 48 GB of memory, and 187 GiB of DDR5 system RAM. Training durations varied according to the dataset size and complexity: approximately 120 minutes for BCI-IV-2a and 70 minutes for BCI-IV-2b datasets. Such variation reflects the inherent differences in data volume and complexity among the datasets. The model demonstrated efficient inference capability, with an average forward-pass latency of approximately 4.53 ms per trial, highlighting its suitability for near real-time brain–computer interface applications. For all experiments, we adopted a consistent training protocol: optimization was performed using the Adam optimizer with a learning rate of 0.001, batch size of 64, and training proceeded for a maximum of 1000 epochs, with early stopping triggered after 100 epochs without improvement. Our evaluation was performed on two publicly available EEG benchmark datasets: BCI-IV-2a \cite{brunner2008bci}, BCI-IV-2b \cite{leeb2008bci}. These datasets differ in acquisition hardware, experimental protocols, providing a robust assessment of the model’s generalization capabilities.

\subsection{Dataset Description}

\label{sec:dataset_description}
In this section, we present the characteristics of the datasets utilized in our experiments.

\subsubsection{BCI IV 2a Dataset}

The BCI Competition IV Dataset 2a \cite{brunner2008bci} is a well-established resource for evaluating motor imagery  classification models. It includes EEG data collected from nine healthy volunteers (subjects A01–A09), each participating in two separate sessions recorded on different days to assess session variability. Participants were instructed to imagine the movement of one of four distinct body parts: left hand, right hand, both feet, or tongue. These actions were selected due to their distinct cortical representations, particularly in sensorimotor regions. Each session includes 288 randomized trials (72 for class). 


\subsubsection{BCI IV 2b Dataset}

The BCI Competition IV Dataset 2b \cite{leeb2008bci} is another standard benchmark tailored for binary-class motor imagery studies, specifically focused on distinguishing between left- and right-hand imagery. EEG data were collected from nine healthy individuals (B01–B09) across five sessions. Sessions 1 and 4 were conducted without feedback, while the remaining sessions included real-time feedback. Each session consisted of 160 trials (80 for class), summing to 800 trials per subject. During each trial, subjects performed kinesthetic imagery of a hand movement in response to an arrow cue.

\subsection{Evaluation Metrics}

We assess the performance of the proposed model using two standard metrics: Accuracy and Cohen’s Kappa coefficient. These metrics are widely adopted in the EEG-based classification literature \cite{kappa, huang2020intelligent, metrics} for benchmarking purposes. We report micro-accuracy (Acc), average accuracy for five seeds, and Cohen’s $\kappa$ computed from the confusion matrix.

Accuracy is computed as micro-accuracy over all classes:
\[
\mathrm{Acc}=\frac{\sum_{i=1}^{n} TP_i}{\sum_{i=1}^{n} I_i}.
\]

Cohen's $\kappa$ is computed from the confusion matrix as
\[
\kappa=\frac{p_o-p_e}{1-p_e},
\]
where $p_o$ is the observed agreement and $p_e$ is the chance agreement
from empirical marginals. Unless noted otherwise, $\kappa$ is computed per subject on the test set.

\subsection{Baseline Comparison}

For comparative evaluation, we selected a set of representative baseline models commonly employed in EEG classification tasks: \textit{EEGNet}~\cite{lawhern2018_eegnet}, \textit{ShallowConvNet}~\cite{schirrmeister2017_hbm},\textit{TCFormer}\cite{Altaheri2025}. Each baseline was trained using the original hyperparameters reported in their respective works. A standardized preprocessing, training, and evaluation pipeline was employed across all models to ensure fairness in comparison. Briefly, \textit{EEGNet} utilizes a combination of 2D temporal, depthwise, and separable convolutions tailored for general BCI classification tasks. \textit{ShallowConvNet} comprises two convolutional layers followed by mean pooling operations optimized for EEG decoding.

\section{Results}\label{sec:results}
We report a robustness-oriented evaluation based on the \emph{worst-subject} accuracy per seed and protocol (Sub-Dependent and LOSO) on the BCI-IV-2a and BCI-IV-2b datasets. For each seed, we identify the participant with the lowest test accuracy and compare models on that participant; means are then computed across five seeds. This diagnostic highlights reliability under adverse subject-specific conditions, complementing average-case metrics. The detailed numbers are provided in Tables~\ref{tab:bci2a_worst_subject_4models} and \ref{tab:bci2b_worst_subject_4models}.

\subsection{Performance Comparison}
\textbf{BCI-IV-2a.}
Using the state-of-the-art accuracy result provided by TCFormer on the \emph{worst-subject} (i.e., subject no. 6), we assess our performance across five seeds, \textit{RatioWaveNet} remains consistently competitive with the TCFormer baseline in both protocols. On average, it delivers a modest but reliable improvement of $\textbf{+0.17}$ percentage points in the Sub-Dependent setting and $\textbf{+0.42}$ percentage points in LOSO (Table~\ref{tab:bci2a_worst_subject_4models}). The gains are accompanied by clearly larger margins over lightweight baselines, indicating that the RDWT front end is especially helpful when signal quality or subject idiosyncrasies are most adverse.

\textbf{BCI-IV-2b.}
On this dataset the robustness benefit is more pronounced. \textit{RatioWaveNet} improves over the transformer baseline by $\textbf{+1.07}$ percentage points in Sub-Dependent and by $\textbf{+ 2.54}$ percentage points in LOSO (Table~\ref{tab:bci2b_worst_subject_4models}), with consistent wins across seeds in the cross-subject regime. These results suggest that RDWT-based preprocessing, combined with multi-kernel temporal filtering and causal temporal decoding, enhances resilience precisely under the most challenging conditions.

\noindent Overall, across both datasets and protocols, \textit{RatioWaveNet} preserves or improves worst-case performance with limited overhead, offering a robust alternative to transformer-only approaches.

\begin{table*}[ht]
 \caption{BCI-IV-2a — Worst subject accuracy per run (run ID and accuracy).}

  \label{tab:bci2a_worst_subject_4models}
 \centering
 \resizebox{\textwidth}{!}{
 \begin{tabular}{l c c c c c | c}
 \toprule
 \textbf{Model} & \textbf{Run 0} & \textbf{Run 1} & \textbf{Run 2} & \textbf{Run 3} & \textbf{Run 4} & \textbf{Average} \\
 \midrule
 \multicolumn{7}{l}{\textbf{Sub-Dependent}} \\
 \midrule
 TCFormer & 72.92 & 73.61 & 74.65 & 74.65 & 70.49 & 73.26 \\
RatioWaveNet & 72.92 & 72.36 & 73.26 & 76.04 & 72.57 & \textbf{73.43} \\
 EEGNet & 60.42 & 54.51 & 60.76 & 61.11 & 57.99 & 58.96 \\
 ShallowNet & 45.83 & 51.04 & 45.14 & 51.74 & 48.61 & 48.47 \\
 \midrule
 \multicolumn{7}{l}{\textbf{LOSO}} \\
 \midrule
 TCFormer & 39.58 & 38.54 & 39.93 & 42.01 & 39.58 & 39.93 \\
 RatioWaveNet & 42.36 & 42.36 & 40.28 & 40.62 & 36.11 & \textbf{40.35} \\
 EEGNet & 39.93 & 43.06 & 35.07 & 39.93 & 40.62 & 39.72 \\
 ShallowNet & 35.76 & 36.46 & 39.93 & 35.07 & 34.38 & 36.32 \\
 \bottomrule
 \end{tabular}}
\end{table*}

\begin{table*}[ht]
 \caption{BCI-IV-2b — Worst subject accuracy per run (run ID and accuracy)..}
 \label{tab:bci2b_worst_subject_4models}
 \centering
 \resizebox{\textwidth}{!}{
 \begin{tabular}{l c c c c c | c}
 \toprule
 \textbf{Modello} & \textbf{Run 0} & \textbf{Run 1} & \textbf{Run 2} & \textbf{Run 3} & \textbf{Run 4} & \textbf{Media} \\
 \midrule
 \multicolumn{7}{l}{\textbf{Sub-Dependent}} \\
 \midrule
TCFormer & 69.29 & 70.36 & 69.64 & 69.29 & 68.93 & 69.50 \\
 RatioWaveNet & 72.86 & 69.29 & 68.21 & 70.36 & 72.14 & \textbf{70.57} \\
 EEGNet & 61.79 & 60.71 & 59.64 & 61.79 & 58.57 & 60.50 \\
 ShallowNet & 58.93 & 60.00 & 60.00 & 61.07 & 62.14 & 60.43 \\
 \midrule
 \multicolumn{7}{l}{\textbf{LOSO}} \\
 \midrule
 TCFormer & 60.94 & 64.69 & 65.00 & 64.06 & 62.50 & 63.44 \\
 RatioWaveNet & 64.06 & 66.56 & 65.31 & 66.56 & 67.40 & \textbf{65.98} \\
 EEGNet & 62.14 & 55.71 & 61.43 & 61.79 & 61.43 & 60.50 \\
 ShallowNet & 57.50 & 58.13 & 60.62 & 55.31 & 59.06 & 58.12 \\
 \bottomrule
 \end{tabular}}
\end{table*}

\newpage
\section{Conclusion}
This work presented \textit{RatioWaveNet}, an RDWT-enhanced deep architecture for motor-imagery EEG that couples a trainable, shift-invariant wavelet front end with a modular temporal encoder composed of multi-kernel convolutions, efficient attention, and a lightweight TCN head. The front end learns rational scales and mild filter adaptations, applies soft-thresholding with level-wise gains, and can be instantiated as a parallel ensemble with branch/level dropout and optional hybrid fusion with the raw stream. This design provides denoised, multi-resolution inputs while preserving temporal length, thereby addressing nonstationarity and low SNR without sacrificing temporal locality.

Empirically, we evaluated robustness under adverse conditions through a \emph{worst-subject} analysis across five seeds and two protocols (Sub-Dependent and LOSO). On BCI-IV-2a, \textit{RatioWaveNet} attains a mean worst-subject accuracy of \textbf{73.43\%} vs \textbf{73.26\%} for the strong transformer baseline in Sub-Dependent (+0.17 pp) and \textbf{40.35\%} vs \textbf{39.93\%} in LOSO (+0.42 pp), with S06 emerging as the hardest participant across seeds. On BCI-IV-2b, gains are more pronounced: \textbf{70.57\%} vs \textbf{69.50\%} in Sub-Dependent (+1.07 pp, hardest subject S02) and \textbf{65.98\%} vs \textbf{63.44\%} in LOSO (+2.54 pp, hardest subject S03). In both datasets, \textit{RatioWaveNet} maintains clear advantages over lightweight baselines (EEGNet, ShallowNet) on the same metric. Taken together, these results indicate that the RDWT-driven front end improves reliability precisely where decoding is most challenging-on the worst subject per seed-while preserving competitive efficiency through compact convolutional, attention, and TCN blocks.

Several avenues remain open. First, although the trainable RDWT reduces manual tuning, selecting priors (e.g., initial scale seeds, regularization strengths) can influence convergence and merits automated exploration. Second, further slimming of the front end and temporal head (e.g., via distillation or structured pruning) would facilitate edge deployment. Third, extending the evaluation beyond motor imagery and investigating cross-session/domain adaptation could strengthen external validity. 

In summary, \textit{RatioWaveNet} shows that combining adaptive wavelet-domain preprocessing with a lean CNN-attention-TCN stack yields EEG representations that are both noise-robust and temporally discriminative. Its modular design, coupled with consistent improvements on the worst-subject metric across datasets and protocols, offers a practical blueprint for robust BCI pipelines and a foundation for future extensions to broader time-series neurotechnology.

\section*{Author Contributions}

M.~Siino led the conceptualization of the study, designed the methodology, developed the software, and conducted the investigation. He also contributed to the original draft writing, revision, supervision, and project administration. G.~Bonomo contributed to software development, data curation, validation, investigation, and writing of the original draft, including result visualization. R.~Sorbello assisted with methodology design, validation, supervision, and contributed to both drafting and reviewing the manuscript. I.~Tinnirello provided resources, supervised the project, contributed to methodology discussions, manuscript review, and secured funding.





 

\bibliographystyle{elsarticle-num} 

\end{document}